# A Novel Approach to OCR using Image Recognition based Classification for Ancient Tamil Inscriptions in Temples

Lalitha Giridhar   Aishwarya Dharani   Velmathi Guruviah,[1]
School of Electronics Engineering (SENSE), VIT, Chennai Campus, Chennai - 600 127.
{lalitha.giridhar2015, aaishwarya.dharani2015, velmathi.g}@vit.ac.in

ABSTRACT

Recognition of ancient Tamil characters has always been a challenge for Epigraphers. This is primarily because the language has evolved over several centuries and the character set has over this time both expanded and diversified. This proposed work focuses on improving optical character recognition techniques for ancient Tamil script which was in use between the 7th and 12th centuries. While comprehensively curating a functional data set for ancient Tamil characters is an arduous task, in this work, a data set has been curated using cropped images of characters found on certain temple inscriptions, specific to this time period as a case study. After using Otsu thresholding method for binarization of the image, a two-dimensional convolution neural network is defined and used to train, classify, and recognize the ancient Tamil characters. To implement the optical character recognition techniques, the neural network is linked to the Tesseract using the Pytesseract library in Python. As an added feature, this work also incorporates Google's text to speech voice engine to produce an audio output of the digitized text. Various samples for both modern and ancient Tamil were collected and passed through the system. It is found that for Tamil inscriptions studied over the considered time period, a combined efficiency (OCR and text to speech) of 77.7% can be achieved.

*Keywords:* Optical Character Recognition; Ancient Tamil script; Convolution Neural Network, TTS

## 1. Introduction

Inscriptions and manuscripts are important sources of information to understand the history and culture of ancient civilizations. Inscriptions are ubiquitous in ancient Indian temples. They can be found on everything - from rocks, slabs, and pillars, to the walls of the temples. Most of these inscriptions convey vast and useful monarchical information about proceedings of administrative and religious processes. These inscriptions are valuable documented proof to understand better the quality of life during the specified era. Tamil Nadu tops the list in the Survey list of Survey of Indian Epigraphy (1996) [1]. This implies that Tamil Nadu has a significantly large number of inscriptions.

Tamil is an ancient language of the world, and amongst the earliest languages in the Indian subcontinent. Since the language has evolved for centuries, it has gone through an exhaustive line of character sets like *Sangakala Tamil, Karoshti, Brahmi, Vat- teluttu, Grandham* and modern character set [2]. To perfect an OCR method for digitizing ancient Tamil inscriptions is therefore technically challenging.

The primary objective of this work is to aid in the development of an Optical Character Recognition (OCR) system to digitize and produce an audible speech output for ancient Tamil inscriptions found on the walls of south Indian temples, built mainly between the 7$^{th}$ and 12$^{th}$ centuries [3]. The inscriptions found on temple walls during this time were inscribed under the Chola dynasty. Many popular names such as Rajaraja Chola-I, Rajendra Chola-I, and Kulothunga Chola-I, served as kings during this period. Under the Chola rule, the Tamil kingdoms had reached new heights of excellence in art, religion, music, and literature [4]. However, most of the inscriptions still remain illegible and untranslated to modern scripts. Of the deciphered inscriptions the contents were often about wage and employment details, and information about rules and regulations to be followed for ceremonial offerings in the temple.

Another notable challenge faced during performing OCR techniques on ancient text relates to the fact that the typeset- ting process was extremely noisy. The hand-carved blocks were uneven and often failed to perfectly sit on the baseline. This poses an issue for segmentation purposes, since it would be hard to eliminate the excessive unnecessary background disturbances and isolate useful information.

[1] Corresponding Author: Dr. G Velmathi, Professor, VIT Chennai Campus, Chennai 600127; velmathi.g@vit.ac.in

The following section provides a survey of relevant literature, while the approach followed is described in Section 3. Results and discussion follow in Section 4.

## 2. OCR for Historical Documents

OCR is the recognition of printed and/or handwritten characters by a computer [5]. OCR has become a widely exploited tool by those interested in digitizing written records. Through the years, significant amount of research has been invested into perfecting and increasing the accuracy of OCR techniques for many global languages such as English, Chinese etc. Despite the funding into OCR research, the world is still relatively far from perfecting the technique for many languages.

Since OCR is a powerful tool, it can be used to digitally immortalize ancient documents and other hard to decipher inscriptions and scriptures from the early centuries. OCR for regional Indian languages still remains a vastly unexplored field with immense potential. Although research work has been carried out in the recent past to perfect character recognition for some ancient Tamil script, 100% accurate results have still not been obtained.

There exist many different tools to perform OCR techniques on text characters. Just recently a system for Tamil Character recognition using the techniques of OCR and Neuro Linguistic Programing was suggested in 2017 [6]. In 2012 a character recognition system using SIFT algorithms for $7^{th}$ century Tamil scripts was designed [7]. A comparative study of Tamil OCR techniques was also carried out in 2009 [8]. In 2016 several key challenges to Tamil character recognition was studied [9].

Many researchers have also worked on identifying characters on inscriptions and performing OCR for them. In 2013, work was carried out [10] for a probabilistic training model for the unsupervised transcription of historical documents.

Seeking inspiration from the work already done, and incorporating a text to speech facility, the current work aims at integrating OCR techniques for ancient Tamil script as well as the more advantageous benefits of an audio output into a single unit.

## 3. Material and Methods

The primary aim of the proposed work is to take a sample inscription from the walls of the ancient temples, scan this image using an OCR scanner designed from scratch to be able to feed it through the CNN [11] linked to the Tesseract [12], thereby using both image recognition techniques and machine learning techniques to produce a digitized equivalent modern-day Tamil translation along with a text to speech application.

### *3.1. Hardware Requirements*

The following mentioned are the hardware requirement used for efficient implementation of the work.

To train the characters from the ancient Tamil inscriptions, high resolution images of some sample inscriptions are scanned, using a scanner. This OCR scanner consists of a bottom plate that is affixed with an LED strip panel that is then covered with a clear glass slab to act as the surface for placing the image as well as a clear pathway for illumination.

The support rods on either side of the box serve two purposes. The first being that the ridges are placed with spacing to fit a wooden slab on which the webcam will be mounted. The second use is that this slab can be adjusted to increase or decrease the level of zoom required to scan the image clearly.

The final component that brings this scanner to life is the webcam that is attached to the slab that captures the image placed on the glass panel and transfers it to the laptop for training the neural network. A Bluetooth speaker that is connected to the laptop is also required to be used to produce the audio output after the text to speech conversion.

### *3.2. Software Implementation*

The following procedure is to be followed to obtain the digitized output of the sample inscription.

i)The input image is first converted to grayscale [13] and the binarization process is applied to the grayscale form of the image. ii) Next, the Otsu's thresholding algorithm [14] is implemented to perform the binarization which involves removing the foreground text from the noisy background.

Otsu's thresholding method involves iterating through all the possible threshold values and calculating a measure of spread for the pixel levels each side of the threshold, i.e., the pixels that either fall in the foreground or the background. The aim is to find the threshold value where the sum of foreground and background spreads is at its minimum. iii) Then, the binarized image obtained is sliced to equivalent letter blocks each containing an ancient Tamil character. iv) Next, this set of cropped images is fed to the convolution neural network; trained for image classification and recognition. To train the blocks of images from the available sample, a data set is curated which is a combination of modern Tamil fonts and ancient fonts (as obtained after binarization). Transfer Learning [15] along with data augmentation is used to train the CNN to classify Tamil letters using Keras [16] and TensorFlow [17]. The advantage of using Transfer Learning comes from the fact that pre-trained networks are adept at identifying lower level features such as edges, lines, and curves.

Since this is often the most computationally time-consuming part of the learning process, using the preset weight values increases the convergence rate significantly.

The processing of an image in a machine happens in terms of numbers. Hence, every pixel in the image is given a value ranging between 0 and 255. This process is known as image encoding [18] and is the primary step for training the neural network to recognize the images.

The encoded image is passed through a convolution layer of dimensions $28 \times 28 \times 1$, where, the 28 represents the size of the image and 1 stands for the number of channels here being binary. Using optimizer techniques in mind, the parameters were updated after every iteration. These layers convolve around the image to detect edges, lines, blobs of colors and other visual elements. For 2D CNN and subsequently two layers of MaxPooling, the dimensions work out to be $32 \times 4 \times 4$. This dimensioning of the CNN has been made after trial and error- keeping in mind the convergence speed as well as the accuracy required. v) Next, the pooling layers reduces the dimensionality of the images by removing a certain number of pixels from the image [19]. To implement this work MaxPooling technique was preferred [20].

The convolved image is compared with every image in the data set. Based on Euclidean distance principle [21], the letter block in the image is recognized to its closest font family.

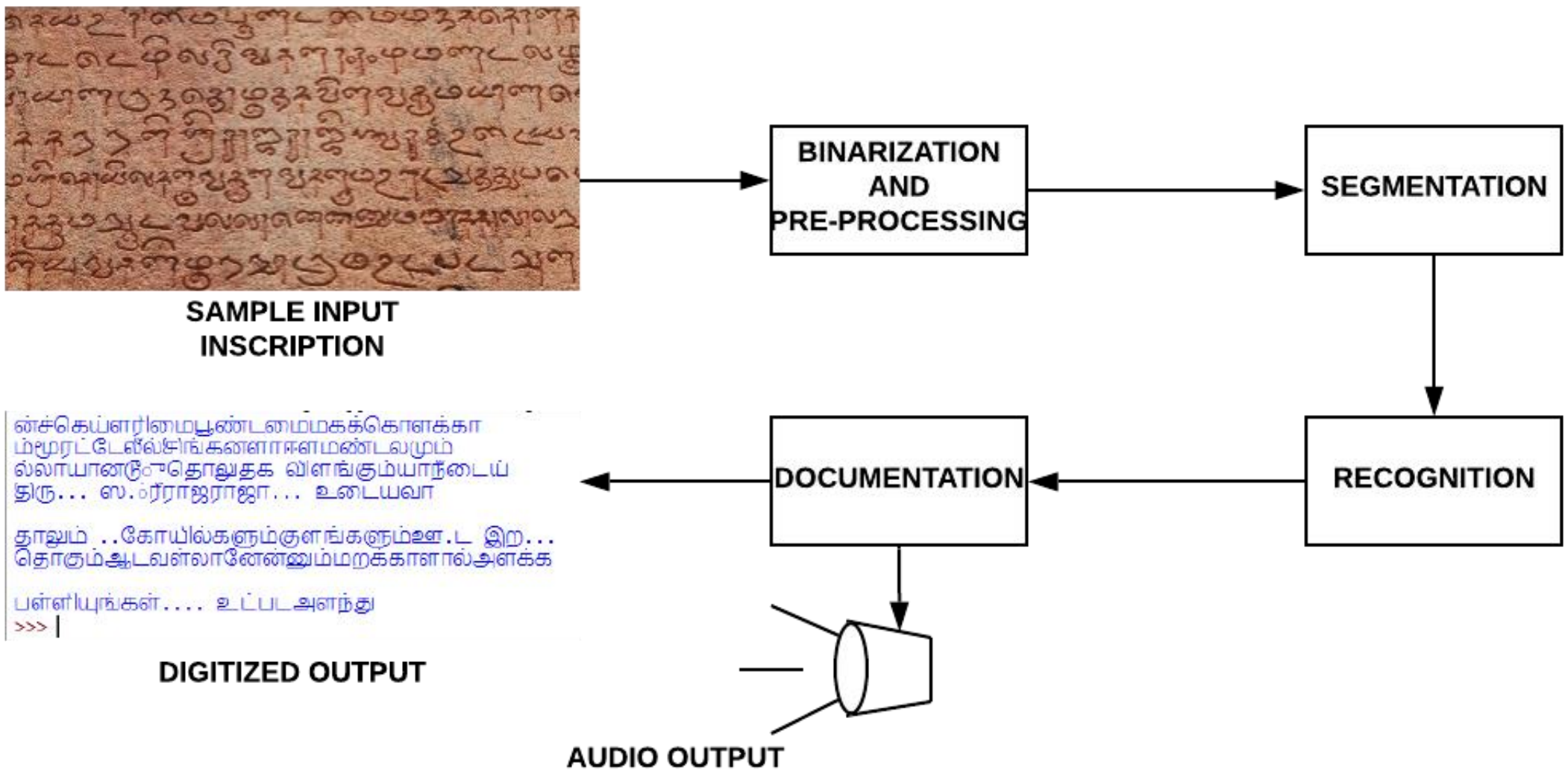

Fig. 1. Proposed architecture of the work

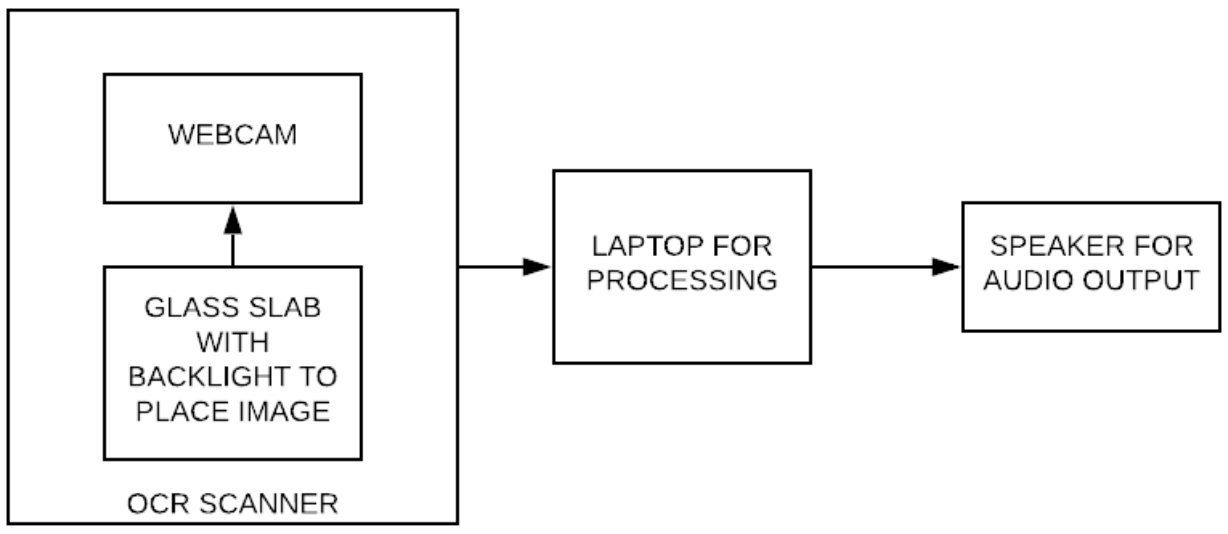

Fig. 2. Suggested setup for implementation

The model is trained to minimize the distance between images belonging to same font family and to maximize the distance between images belonging to different font family. The recognized images of each letter block are combined to form a single image tile. This is then subjected to father processing to perform OCR.

To implement OCR, the tile is fed to the Pytesseract library [22] trained for Tamil. This inbuilt library of python performs the OCR techniques on the image tile again to produce the digitized output text in a readable and editable form. This step is performed to double validate the results of the image recognition performed by the neural network training model, thereby making it a necessary step to obtain accurate results.

vi) Finally, for the conversion of the digitized text to an audio output, the "gTTS" library [23] of python is used. gTTS is a tool to interface with Google Translate's text-to-speech API [24], which supports automatic retrieval of wide number of languages. This attribute is exploited to translate the text in Google engine's Tamil voice [25]. This easy and feasible text to speech feature integrated with the digitized text output is the expected and highlighted output.

## 4. Results and Discussion

This work aims on developing an integrated system for digitization and audio documentation of ancient Tamil script. Many scanned sample images of inscriptions from various ancient Tamil temples were obtained curated and tested using the developed methodology.

The observed results were documented. Since there is no standardized data set available for ancient Tamil scripts, in this work a minimalistic data set has been made use of which would suffice for a specific inscription. Using the image slicer tool from Python [26], the cropped characters were fed through a two-dimensional CNN to train and produce the equivalent digitized text output on the python shell screen and its equivalent audio output on the speaker.

### *4.1 Modern Tamil script*

We tested the efficiency of the developed OCR with integrated text to speech system for modern Tamil script, by using many samples were obtained from different sources. Accurate modern equivalent reproduction of a character was considered a success. Incorrect character recognition, or unrecognized characters were considered as faulty output. Some of the tested samples included partially pre-digitized text, text from printed Tamil literature, and handwritten text.

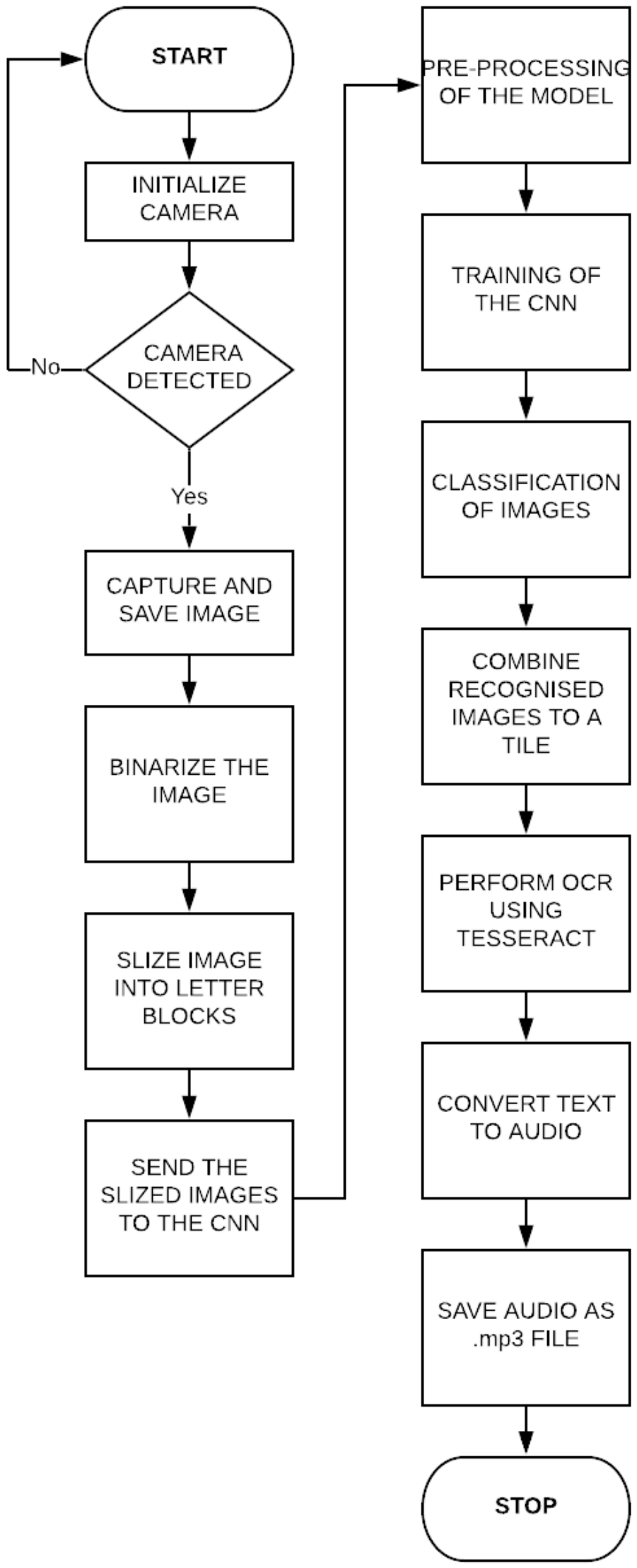

Fig. 3. Flowchart for software implementation

In the case of the pre-digitized text, the OCR accuracy was close to 99%. The real challenge was to perfect the audio output. It was observed that the audio output after performing text to speech conversion also worked with near 99% accuracy. Since the accuracy of the audio output is dependent on the accuracy of the OCR, the efficiency of the integrated system relies primarily on the efficiency of the OCR technique.

For testing the system on printed Tamil literature, some popular regional books were used as samples. High quality images from these books were then scanned using the developed OCR scanner and tested. The accuracy of the system for printed Tamil literature averaged around 91%. The difficulties that were faced during the testing lies in the segmentation stage of the process. Due to the bend on the spine of the book, some images taken for scanning and pre-processing failed to be accurately segmented.

The error in segmentation led to certain characters being incorrectly recognized. Consequently, in the case of text to speech implementation, the efficiency significantly dropped to 87%. This is because incorrect character recognition in the OCR part of the system leads to incorrect word recognition in the case of text to speech.

Finally, certain handwritten samples were obtained from volunteers to be tested using the developed system. In the case of the handwritten samples, OCR efficiency strictly relied on proper segmentation of the written words. In free hand writing of modern Tamil script, often perfect segmentation is not accounted for. Since a threshold-based method is used to determine the right level of segmentation in this work, segmentation accuracy varied drastically between different samples. Another point to be taken into consideration would be that the training samples used for training the CNN were a digitized character set. Due to this, handwritten characters were mapped for identification alongside digitized samples. The accuracy of OCR system for handwritten modern Tamil script was therefore around 70%. The text to speech accuracy was 68%.

It is important to note that the reason for testing the system for its accuracy in modern Tamil script was only to set a baseline for the ancient Tamil script. Ancient inscriptions are a form of handwritten text due to their lack of uniformity between different occurrences of a single character. Therefore, training the CNN to for modern Tamil handwritten script, enabled overcome the challenge of un-uniformity amongst ancient text.

### *4.2 Ancient Tamil Script*

To test the accuracy of the system on ancient Tamil scripts, samples were taken from some historical temples in south India. The high-quality images of the inscriptions were then scanned using the OCR Scanner and subsequently fed as input images to test the working of the developed system. Since the dataset for identification of ancient Tamil script characters is not as extensive as that available for modern script, an alternative approach was used to train the characters. Using ImageSlicer [26] tool from the python library, the individual characters of each sample image tested were identified and cropped to curate a dataset. This enabled higher image recognition percentge of the characters after training using the convolution neural network. From the thirty plus samples that were tested, it was observed that the efficiency of the system in successfully performing OCR and text to speech conversion for the input images varied from sample to sample. The accuracy for certain specific samples tested are tabulated in Fig 6. The varying nature of the efficiency of the system can be attributed to three reasons.

Firstly, since the available dataset is sparse, and a universal solution is still at its early stages. However, upon isolation of a specific case, it is possible to use machine learning techniques to efficiently train the dataset of characters available in the image. As the Tamil language has also evolved over the decades, it is increasingly hard to obtain a standardized list of characters for training ancient characters. This caused certain characters to be wrongly identified by the system. As mentioned before, like handwritten scripts, scripts found on inscriptions are not uniform and this also causes recognition problems.

The second reason why accuracy is affected lies in inability of the sample image to be accurately binarized. Since the inscriptions are not of uniform spacing and the base (i.e., rocks, pillars, etc.) on which they have been inscribed often an uneven surface, errors occur during binarization stage. The erosion on these surfaces further lead to certain characters remaining unrecognizable or lost during binarization. These errors could be minimized by using images of higher resolution, and perhaps also digitally minimizing the background erosion before binarization. Protection of inscriptions from further degradation is also vital to aid in this process and to preserve the historical wonders.

Lastly, since a language parser [27] has not been optimized for ancient Tamil, segmentation of words from the digitized output obtained remains out of scope for the current work. However, upon consultation with experts, manual segmentation techniques can be applied to provide a further depth to understanding the inscriptions.

A successful OCR conversion is determined by identification of individual character block followed by correct recognition of modern Tamil character equivalent and accurate reproduction of character in the final output.

The combined efficiency of the system can be found to be 77.7%. This percentage efficiency of the system was computed by averaging the individual efficiency rate for each sample. It is important to acknowledge the realistic constraints encountered while perfecting this system and although the efficiency is oscillatory in nature, there is a definitive scope to achieve higher constant efficiency rates soon upon overcoming the discussed challenges.

## 5. Conclusion

This work attempted at creating a universally applicable OCR system integrated with audio output for ancient Tamil script. By using a CNN and Image Recognition techniques, an operable system was designed for modern and ancient Tamil. The difference in the style of the ancient Tamil scripts from the modern Tamil script posed as a challenge to execute the task efficiently. Multiple samples of ancient inscriptions from some historical temples were taken as case studies to implement the developed methodology. An acceptable accuracy rate of 77.7% was attained for these samples. The outputs obtained could not be digitally segmented due to lack of availability of any language parser for ancient Tamil scripts. OCR techniques for ancient Tamil scripts is a rich research topic. The efficiency of OCR techniques on ancient Tamil scripts can be improved by expanding the available data set. Another possible extension of our work is to develop a language parser to aid in segmentation of the digitized script and to improve the accuracy of the audio output. There is also scope for implementation on a broader scale with near total efficiency with the help of a more thoroughly researched data set, not only for the Tamil language, but also for several other culturally rich languages of India.

Table 1. Results Obtained

| | Scripture/document/temple inscription | Grayscale Output | Binarized Output | Digitized text Output in Python shell |
|---|---|---|---|---|
| **Case 1: Modern Tamil** | 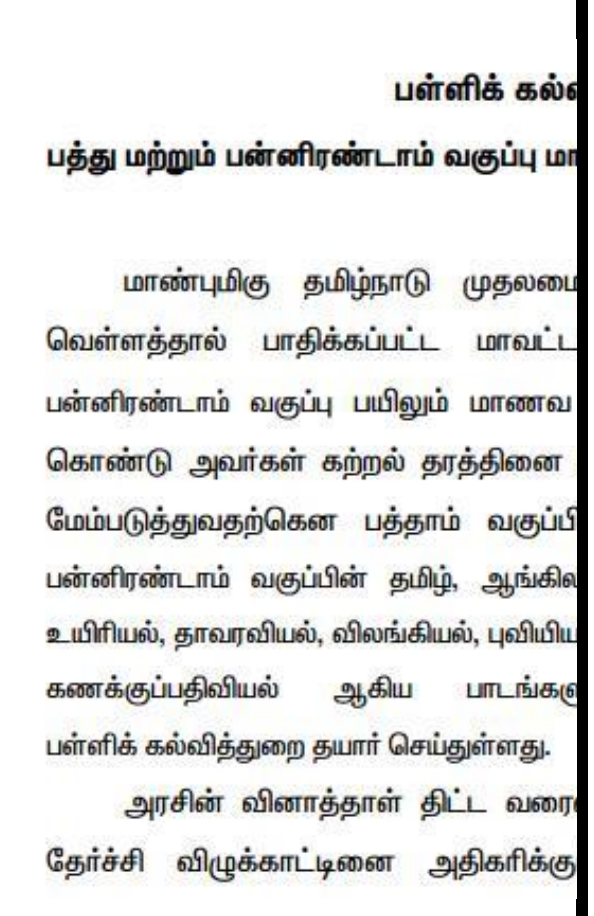 | 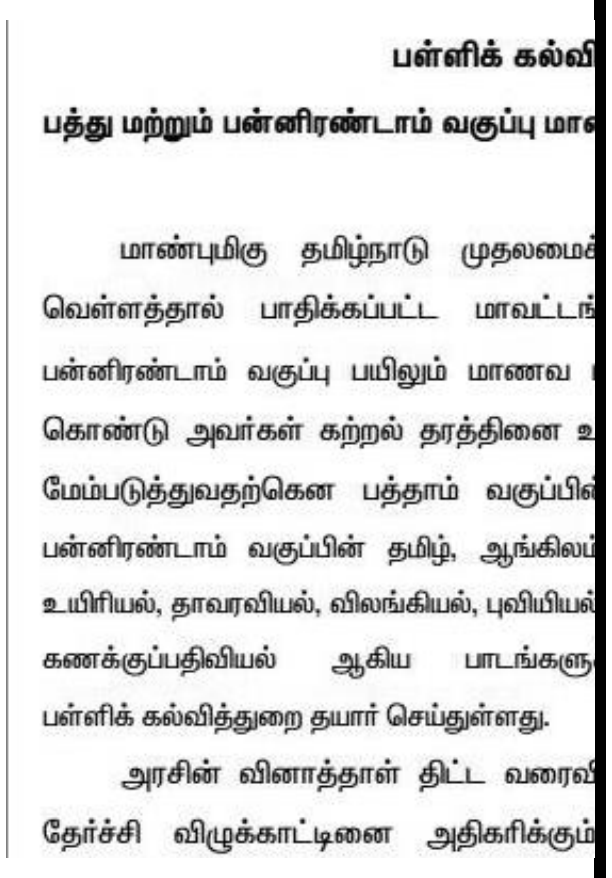 | 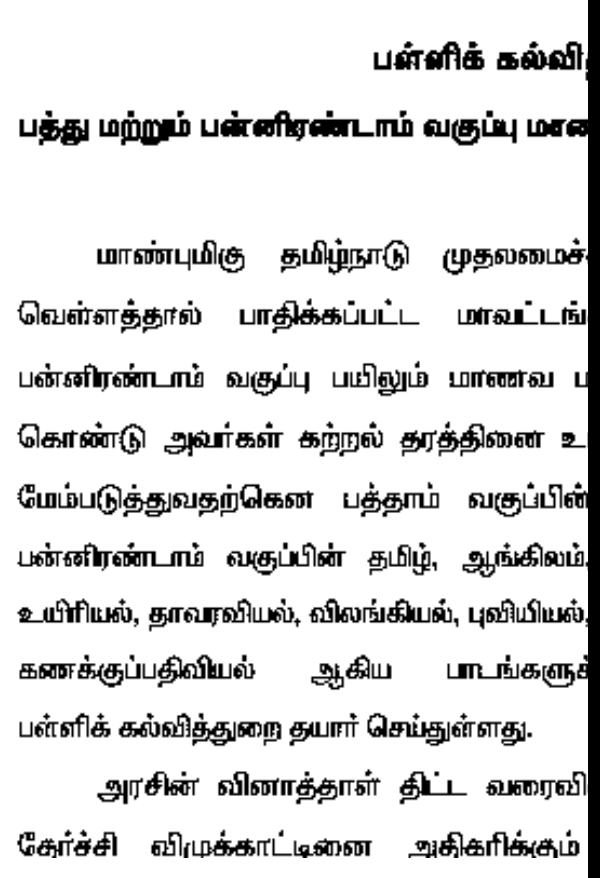 | பள்ளிக் கல்வித்துறை<br>பத்து மற்றும் பன்னிரண்டாம் வகுப்பு<br><br>மாண்புமிகு தமிழ்நாடு முதலமைச்சர்<br>வெள்ளத்தால் பாதிக்கப்பட்ட மாவட்ட<br>பன்னிரண்டாம் வகுப்பு பயிலும் மாண<br>கொண்டு அவர்கள் கற்றல் தரத்தினை<br>மேம்படுத்துவதற்கென பத்தாம் வகுப்<br>பன்னிரண்டாம் வகுப்பின் தமிழ், ஆங்<br>ல்,<br>உயிரியல், தாவரவியல், விலங்கியல்,<br>கவியல்,<br>கணக்குப்பதிவியல் ஆகிய பாடங்களுக்<br>பள்ளிக் கல்வித்துறை தயார் செய்துள்<br><br>அரசின் வினாத்தாள் திட்ட வரைவின்<br>தேர்ச்சி விழுக்காட்டினை அதிகரிக்கும் |
| **Case 2: Ancient Tamil** | 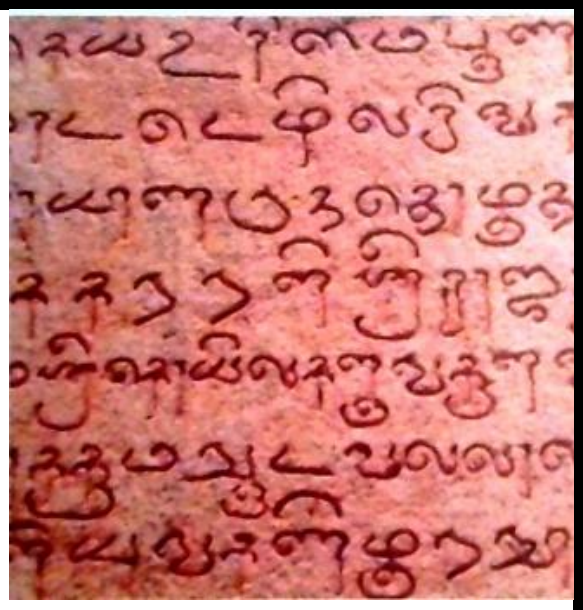 | 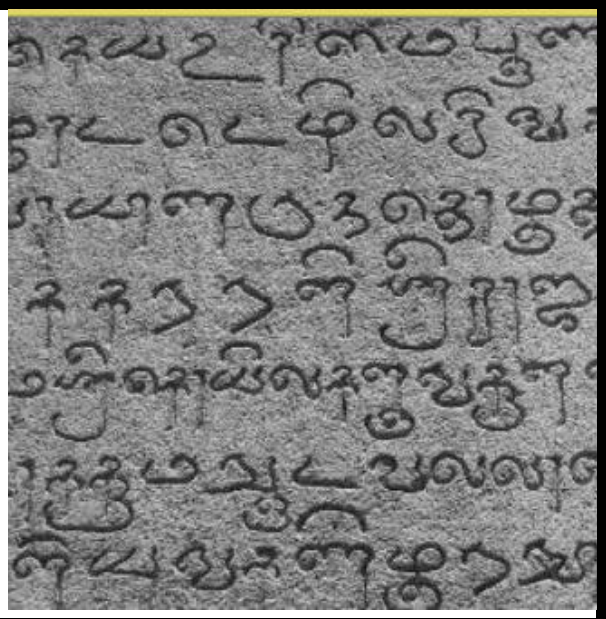 | 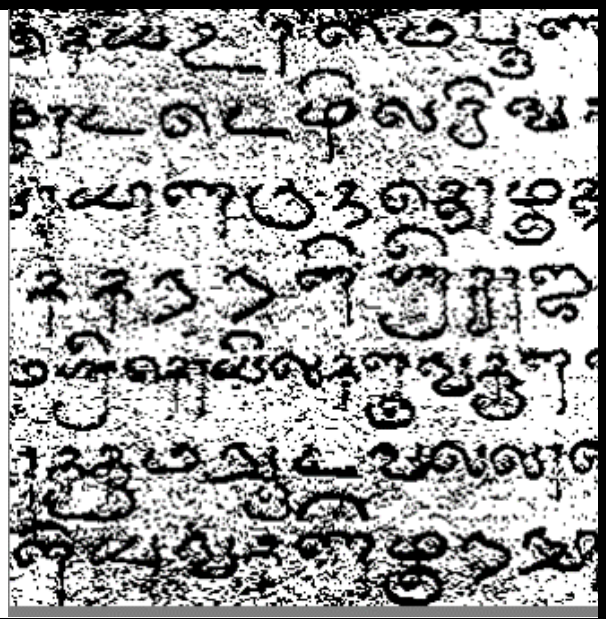 | ன்ச்கெய்ளரிமைபூண்டமைமகச<br>ம்ஹூரட்டேலீல்சிங்கனளாஈளமன<br>ல்லாயானடுெதொலுதக விளங<br>திரு... ஸ.ர்ரீராஜராஜா... உன<br><br>தாலும் ..கோயில்களுங்குளங்<br>தொகும்ஆடவள்லானேன்னும்ம<br><br>பள்ளியுங்கள்.... உட்படஅளந<br>>>> |
| **Case 3: Ancient Tamil** | 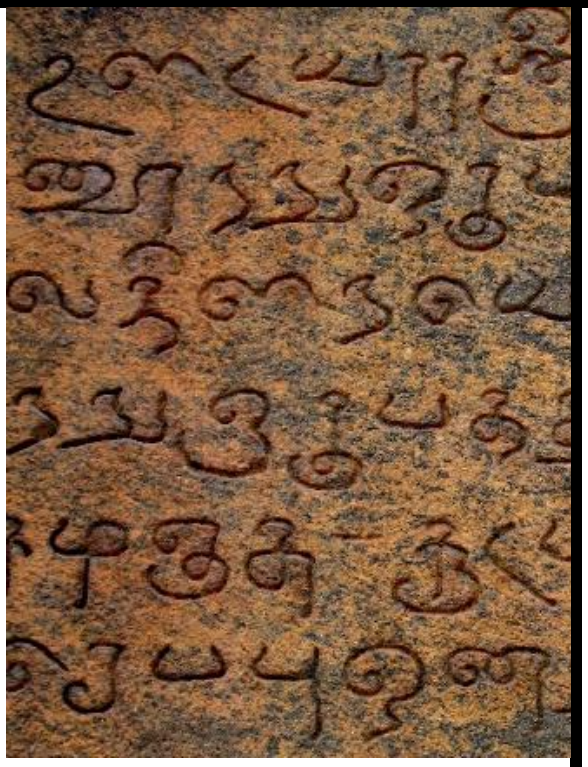 | 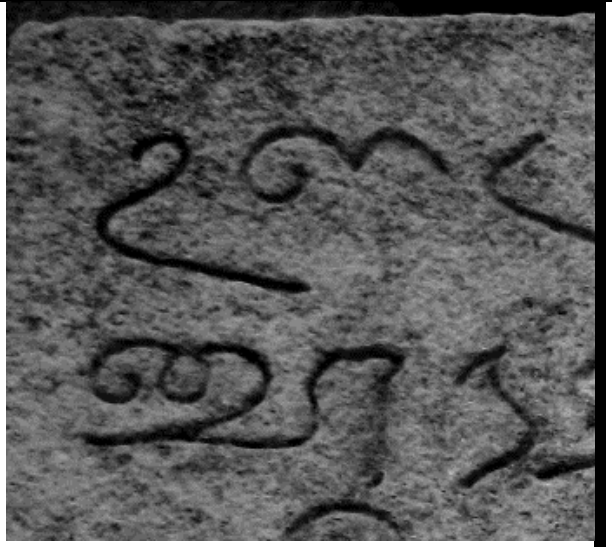 | 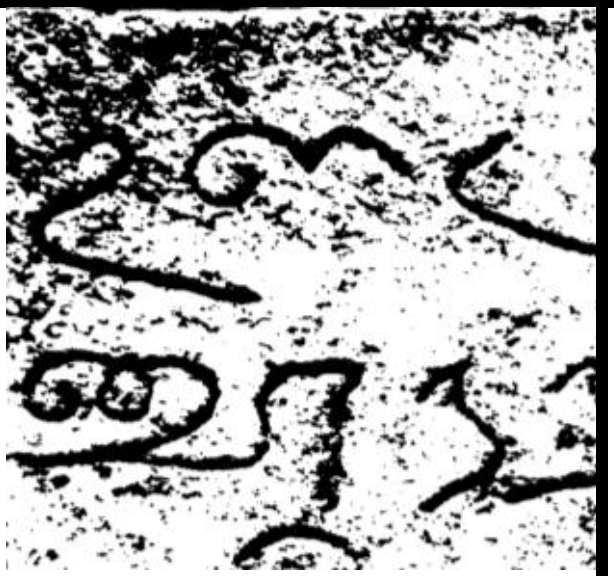 | உடையார்ஸ்ரீராஜ<br>நூராாவஓபதத்தோ<br>ல்கரையெ...கல<br>நவஓபத்துஓமஓ..<br><br>கமஓக குடம்ஓபெ<br><br>ரெபபூஓணா..யுப |
| **Case 4: Ancient Tamil** | 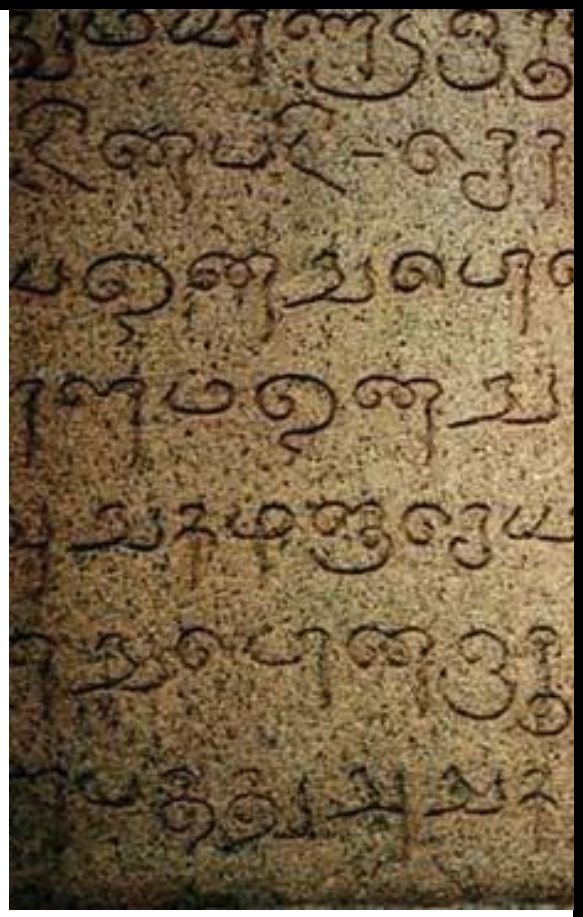 | 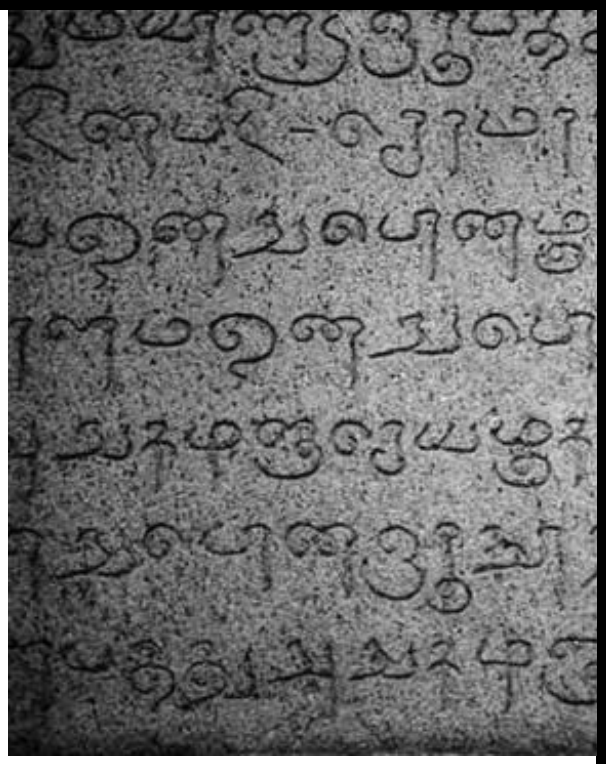 | 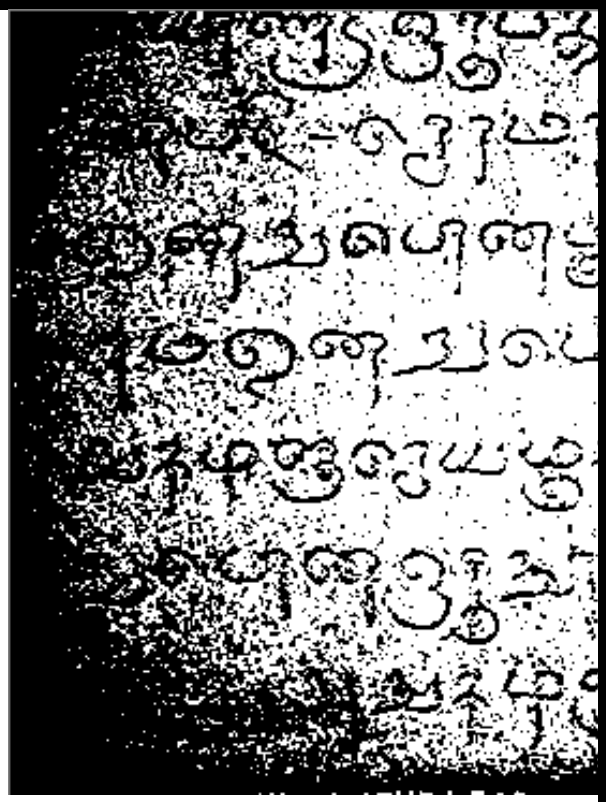 | மயாநேராறுப்ததிதிதாஜவ<br><br>பீத.ரொமாணயெயும்பந்<br><br><br>பாணேய்வாப்போனமுபப<br>மியே நாவுபோண்ணுஒருத<br><br>வாகம்.ரேயமூகக்காலெ<br><br><br>வபோண்ணுஒருத்சாறவத<br><br>பாத்துதுவுவாக்கமு ஒரெப |

| | Inscription | OCR Scanner output for Ancient Tamil | Output Efficiency (audio+text) |
|---|---|---|---|
| **Brihadeeshwara temple, Thanjavur, Tamil Nadu, India** | 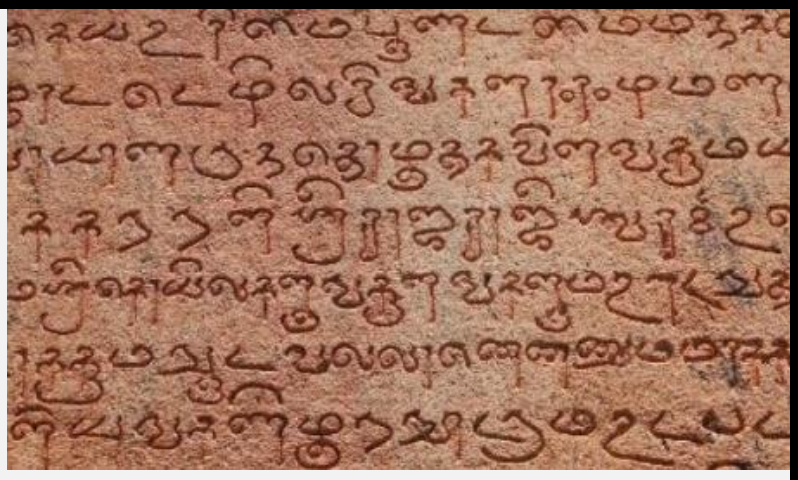 | ன்ச்கெய்ளர்மைபூண்டமைம்மகக்கொளக்கா<br>ம்மூரட்டேலீல்சிங்கனளாஈளமண்டலமும்<br>ல்லாயானடுொதொலுதக விளங்கும்யாந்டைய்<br>திரு... ஸ.ர்ராஜராஜா... உடையவா<br>தாலும் ..கோயில்களும்குளங்களும்ஊ.ட இற...<br>தொகும்ஆடவள்லானேன்னும்மறக்காளால்அளக்க<br>பள்ளியுங்கள்.... உட்படஅளந்து<br>>>> \| | 79.1% |
| **Brihadeeshwara temple, Thanjavur, Tamil Nadu,India** | 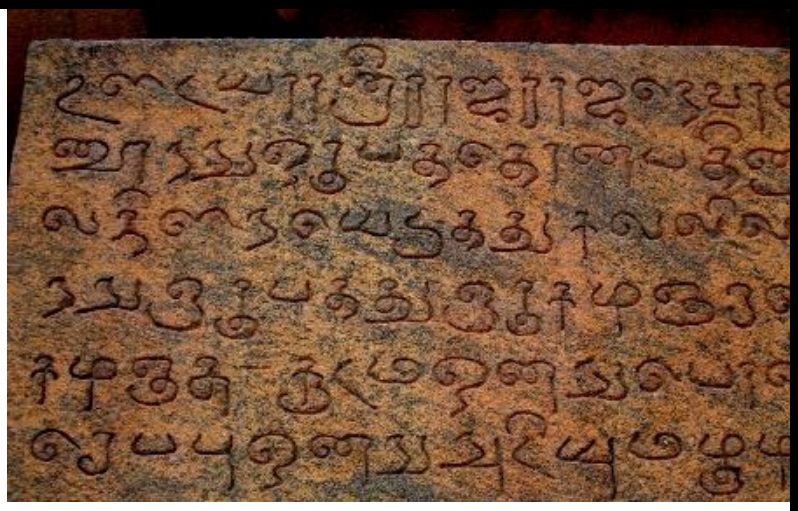 | உடையார்ஸ்ரீராஜராஜ..வரொ<br>நூராவஓபதத்தோன்பதினால்<br>ல்கரையெ...கல்லீ...<br>நவஓபத்துஓமஓ..<br>கமஓக குடம்ஓபொ.மூ<br>ரெபபூஓணா..யுமமூலூ | 79.7% |
| **Brihadeeshwara temple, Thanjavur, Tamil Nadu,India** | 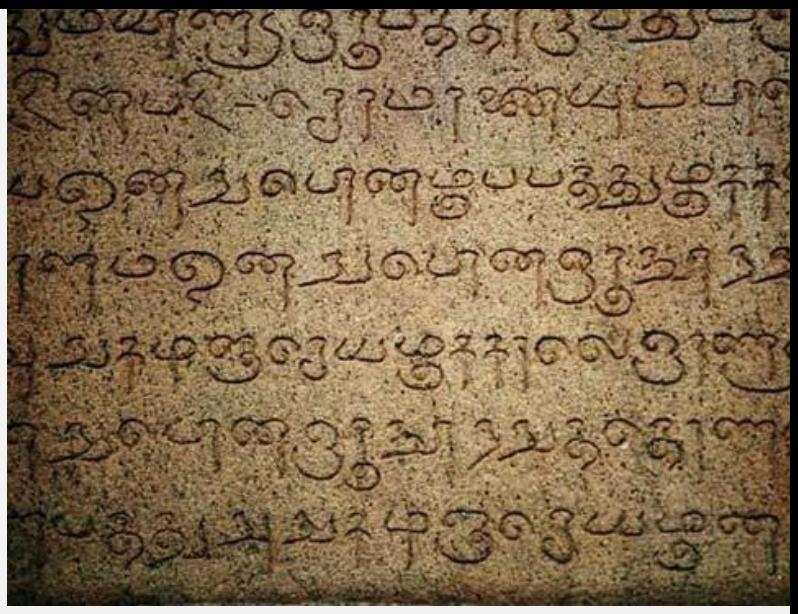 | மயாநேராறுப்ததிதிதாஜஹப்டு.<br>பீத.ரொமாணயெயும்பந்திய<br>பாணேய்வாப்போனமூபப்பட்இத்மாமுக்கமா.<br>ம்யே நாவுபோண்ணுஒருதது. வாத:<br>வாகம்.ரேயமூகக்காலெயோ.மா.<br>வபோண்ணுஒருத்சாறவதத்மொண்ணூறு<br>பாத்துதுவுவாக்கமு ஒரெய் யமூனாஉம் | 81.3% |
| **Brihadeeshwara temple, Thanjavur, Tamil Nadu,India** | 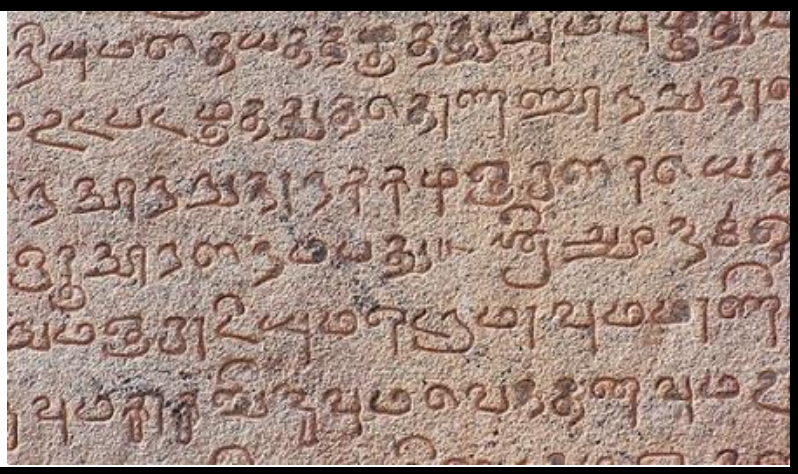 | த்யும்இியெயாத்மூதுவும்பூ<br>உடபடமுததுதோறணை...வாக்களி.<br>ரா.ரவாககராகககமோறயேயததலு.<br>ஒருவிராதயெமபபது ஸ்ரீ....வா.<br>வாமோதிறாதாயும்மாஜேதறமாவுமமான்க்றம்<br>உம்... உம்வெக்தனவுமுஉடபட | 75% |
| **Kasi Viswanathar Temple, Tenkasi, Tamil Nadu, India** | 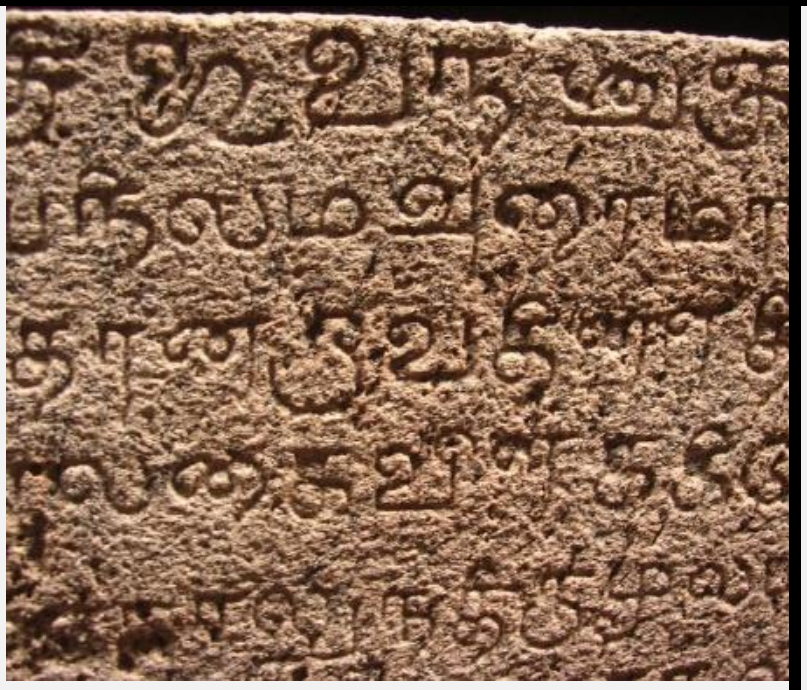 | தோரவனனுராகவ<br>பனல்மஷர்யேமார<br>தரணோரவாத்தாயெத்ர.<br>எடுவ<br>தனாதருமூலப: | 71.1% |
| **Airavatesvara temple, Kumbakonam, Tamil Nadu, India** | 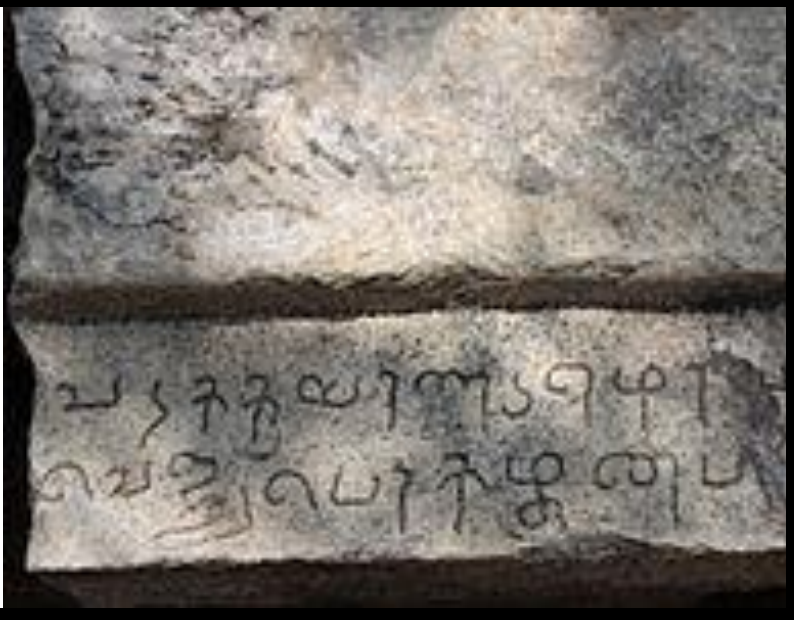 | வடக்கு யானஅமா.<br>வேஜுபோஹறுமூநம் பாண | 80% |